\documentclass[preprint,12pt]{elsarticle}

\usepackage{setspace}
\usepackage[utf8]{inputenc} %unicode support

\usepackage[table]{xcolor}
\usepackage{subcaption}
\usepackage{balance}
\usepackage{amsmath,amssymb}
\usepackage{url}
\journal{Information Sciences}

\usepackage{color}
\usepackage{xcolor}

\usepackage{xspace}
\usepackage{graphicx}
\usepackage{longtable}
\usepackage{caption}
\usepackage{amsmath,amsthm}
\usepackage{multirow}
\usepackage{multicol}
\usepackage{booktabs}
\usepackage{url}
\usepackage{pifont}
\usepackage{color}
\usepackage{supertabular}
\usepackage[export]{adjustbox}

\usepackage{algorithm}
\usepackage{algorithmicx}
\usepackage{algpseudocode}

\usepackage{amsmath}
\usepackage{amsthm,amsmath,amssymb}
\usepackage{mathrsfs}
\usepackage{bm}
\usepackage{subcaption}
\usepackage{hyperref}
\usepackage{CJKutf8}

\usepackage{cleveref}

\floatname{algorithm}{Algorithm}

\newcommand{\vct}[1]{\ensuremath{\boldsymbol{#1}}}
\newcommand{\mat}[1]{\ensuremath{\mathbf{#1}}}
\newcommand{\set}[1]{\ensuremath{\mathcal{#1}}}

\newcommand{\T}{\ensuremath{\top}}

\newcommand{\mynewcomment}[1]{\textcolor{black}{#1}}

\newcommand{\sign}{\text{sign}}
\newcommand{\argmin}{\operatornamewithlimits{\arg\,\min}}

\newcommand{\myparagraph}[1]{\smallskip \noindent \textbf{#1}}
\newcommand{\ie}{{i.e.}\xspace}
\newcommand{\eg}{{e.g.}\xspace}

\newcommand{\imagenet}{{ImageNet}\xspace}
\newcommand{\chmnist}{{HCL2000}\xspace}
\newcommand{\mnist}{{MNIST}\xspace}
\newcommand{\cifar}{{CIFAR-10}\xspace}

\begin{document}

\begin{frontmatter}

\title{Why Adversarial Reprogramming Works, When It Fails, and How to Tell the Difference}

\author[NPU]{Yang Zheng}

\author[NPU]{Xiaoyi Feng}

\author[NPU]{Zhaoqiang Xia}

\author[NPU]{Xiaoyue Jiang}

\author[unica]{Ambra Demontis\corref{mycorrespondingauthor}}
\cortext[mycorrespondingauthor]{Corresponding author}
\ead{ambra.demontis@unica.it}

\author[unica]{Maura Pintor}

\author[unica]{Battista Biggio}

\author[NPU,unige]{Fabio Roli}

\address[NPU]{Northwestern Polytechnical University, Xi’an, China}
\address[unica]{University of Cagliari, Italy}
\address[unige]{University of Genoa, Italy}
%\address[pluribus]{Pluribus One, Italy}


% max 150 word
\begin{abstract}
Adversarial reprogramming allows repurposing a machine-learning model to perform a different task. For example, a model trained to recognize animals can be reprogrammed to recognize digits by embedding an adversarial program in the digit images provided as input. 
Recent work has shown that adversarial reprogramming may not only be used to abuse machine-learning models provided as a service, but also beneficially, to improve transfer learning when training data is scarce. However, the factors affecting its success are still largely unexplained. In this work, we develop a first-order linear model of adversarial reprogramming to show that its success inherently depends on the size of the average input gradient, which grows when input gradients are more aligned, and when inputs have higher dimensionality. The results of our experimental analysis, involving fourteen distinct reprogramming tasks, show that the above factors are correlated with the success and the failure of adversarial reprogramming.

%%%%%%%%%%%%%%%%%%%%%%%%%%%%%%%%%%%%%%%%%%%

\end{abstract}

\end{frontmatter}

%\linenumbers

%neurocomputing no limiti pagine

\section{Introduction}
\begin{figure*}[t]
    \centering
    \includegraphics[width=0.99\textwidth]{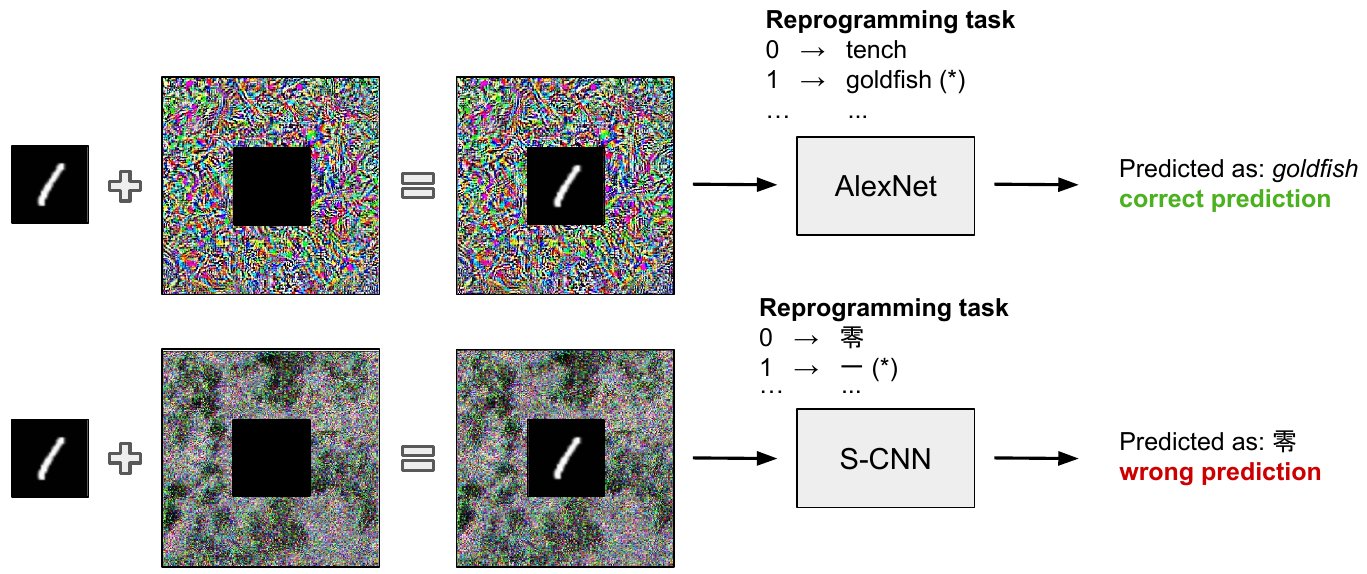}
    \caption{Adversarial reprogramming of AlexNet~\cite{alex12-nips}, trained on the \imagenet dataset (\textit{top}), and CWNet~\cite{carlini17-sp}, trained on Chinese digits (\textit{bottom}), to recognize MNIST handwritten digits. Each digit class is mapped to a different class among those predicted by the target model (\eg, for AlexNet, 0 to \textit{tench},  1 to \textit{goldish}, etc.).
    In the example, the handwritten digit 1 is embedded in both adversarial programs expected to be classified as  \textit{goldfish} by AlexNet and as the Chinese digit 1 by CWNet. However, while AlexNet can be successfully reprogrammed, reprogramming fails for CWNet.}
    \label{fig:repr-success-and-failure}
\end{figure*}

Adversarial reprogramming is a technique that repurposes a machine learning model, originally trained for a task, to perform a different chosen task, without retraining or fine-tuning it. This technique optimizes an adversarial perturbation (\textit{adversarial program}) that can be applied to the model's inputs to make the model perform the chosen task. For example, a model trained to recognize certain classes of samples from a \textit{source} domain (\eg, \imagenet objects) can be reprogrammed to classify samples belonging to a different, \textit{target} domain (\eg, MNIST handwritten digits).
To this end, one should first establish a mapping function between the class labels of the source domain and those of the target domain (\eg, the handwritten digit ``0'' can be associated to the \imagenet class ``tench'', the handwritten digit ``1'' to the \imagenet class ``goldfish'', etc.). 
Once such a class mapping is established, all the target-domain samples are modified to embed the adversarial program, \ie, an adversarial perturbation equal for all the samples (\textit{universal}) optimized against the target model to have the samples of the target domain assigned to the desired source-domain classes. 
An example of adversarial program for reprogramming an \imagenet model to classify handwritten digits is shown in Fig.~\ref{fig:repr-success-and-failure} (\textit{top}). In this case, the adversarial program consists of a frame surrounding the input image, but such programs, as well as other adversarial perturbations, can also be optimized to be superimposed on the input samples.
While reprogramming was initially proposed for images, its application on other domains, such as text~\cite{neekhara2019adversarial} and audio~\cite{yen2021study} classification, is also possible.
Additionally, reprogramming was also shown to be useful for repurposing models across different domains~\cite{yang2021voice2series,neekhara21-arxiv}. For example the authors of~\cite{neekhara21-arxiv} have shown that a model trained to perform classification on the ImageNet dataset can be reprogrammed to perform sentiment analysis and topic classification.

Adversarial reprogramming was originally introduced to abuse machine-learning models provided as a service and steal computational resources by repurposing a model to perform a task chosen by the attacker. For instance, an online service that uses deep networks originally trained to classify images of objects can be reprogrammed to recognize numbers and digits for solving CAPTCHAs\footnote{Completely Automated Public Turing test to tell Computers and Humans Apart.} and automating the creation of fake accounts~\cite{elsayed19-ICLR}. However, it has been recently shown that adversarial reprogramming can also be used for beneficial tasks like transfer learning, in scenarios with scarce data and constrained resources. The authors of~\cite{tsai20-pmlr} have empirically demonstrated that in such scenarios, adversarial reprogramming achieves better performance than fine-tuning, \ie, the classical approach to transfer learning in deep networks. We conjecture this is because, in scenarios with scarce data, the number of parameters that should be optimized becomes of paramount importance. If the parameters are too many, having a big dataset is necessary to learn the task, whereas a small set of parameters can be optimized even with small datasets. Fine-tuning requires modifying the model parameters, which are often a considerable number for deep neural networks. Reprogramming, instead, requires optimizing the parameters of the adversarial program, which are usually a number quite smaller.
Another advantage of reprogramming over fine-tuning is that, unlike fine-tuning, reprogramming allows leaving the model as it is, thus preserving the original functionality for which it was trained. Furthermore, it can also be performed having only black-box access to the machine-learning model (without knowing its internal details, including its architecture and learned parameters)~\cite{tsai20-pmlr}. In this case, one can compute the reprogramming mask by repetitively querying the model and observing its predictions for different inputs. 

Notwithstanding the practical relevance of adversarial reprogramming, the factors affecting its success are still unclear. Even the most recently published papers have not answered the following key questions: when and why adversarial reprogramming works, but most importantly, when and why it fails. 
In the original work that proposed adversarial reprogramming~\cite{elsayed19-ICLR}, the authors successfully applied this technique to solve a complex task, \ie, reprogramming an \imagenet model to classify MNIST handwritten digits. 
However, they also showed that reprogramming did not successfully work when trying to reprogram untrained networks to solve the same task.
While starting investigating the reasons behind such failures, we tried to apply reprogramming to what we thought was a simpler task: reprogramming a small convolutional neural network (CWNet~\cite{carlini17-sp}, described in Table~\ref{tab:small-dnn-architecture}), trained to recognize handwritten Chinese digits, so that it could recognize the MNIST handwritten (Arabic) digits, as depicted in Fig.~\ref{fig:repr-success-and-failure} (\textit{bottom}).
Despite our efforts in tuning the hyperparameters, we found that adversarial reprogramming surprisingly fails on this intuitively simpler task.
Our conjecture was based on the fact that both the source and the target domain in this case are digit recognition problems. However, as we will show later, task similarity does not help adversarial reprogramming.

%(3)
To shed light on the underlying factors affecting success and failure of adversarial reprogramming, in this work we provide a first-order linear analysis of adversarial reprogramming, starting from the observation that adversarial programs can be indeed considered as universal adversarial perturbations (Sect.~\ref{sec:math-tool}). Our analysis shows that the success of reprogramming is inherently dependent on the size of the \textit{average} input gradient, which grows (i) when the input gradients (\ie, the gradients of the classification loss w.r.t. the input values, computed on different target-domain samples) are more aligned, and (ii) when the number of input dimensions increases.
%(4)
To validate the proposed mathematical model, we carry out an extensive empirical analysis (Sect.~\ref{sec:experiments}) involving three different neural-network models and four datasets, resulting in fourteen distinct reprogramming tasks.
Our experimental analysis shows that our first-order model of adversarial reprogramming correctly highlights the main factors behind its success and failure.
%(5)
We conclude the paper by discussing related work (Sect.~\ref{sec:related}), along with the main contributions, limitations and future developments of our work (Sect.~\ref{sec:conclusions}).

\section{Understanding Adversarial Reprogramming}
\label{sec:math-tool}

In this section we develop a first-order mathematical model of adversarial reprogramming, which aims to highlight the main factors underlying its success and failure. To this end, we first formalize adversarial reprogramming as a loss minimization problem (Sect.~\ref{sec:problem-formulation}). We then show that, under linearization of the loss function, the optimal solution can be computed in closed form, and the loss reduction is proportional to the size of the average input gradient (Sect.~\ref{sec:firs-order-analysis}). This in turn means that reprogramming a target model is easier when the input gradients are well aligned, and inputs are high dimensional. We conclude the section by discussing how our analysis can be extended to adversarial programs more general than those depicted in Fig.~\ref{fig:repr-success-and-failure} (Sect.~\ref{sect:extension}).

\subsection{Problem Formulation}
\label{sec:problem-formulation}

Given a model $f$, trained on a source-domain dataset (\eg \imagenet) $\set S$, the goal of adversarial reprogramming is to find a single universal perturbation $\delta$, that can be applied to all the samples of the target-domain data (\eg the MNIST  handwritten digits set), to make the model perform the desired task (\eg classify arabic numbers despite being trained to classify objects).

Let us assume that we have a source-domain dataset $\set S=(\tilde{\vct x}_j, \tilde{y}_j)_{j=1}^m$ and a target-domain dataset $\set T = ( \vct x_i, y_i)_{i=1}^n$, consisting of $m$ and $n$ samples, respectively, along with their labels. In both cases, the input samples are represented as $d$-dimensional vectors in $\set X = [-1,1]^d$, while the class labels belong to different sets, respectively, $\tilde{y} \in \set{\tilde Y}$ and $y \in \set Y$ for the source and the target domain. 
We are also given a target model $f : \set X \mapsto \set{\tilde Y}$ which makes predictions on the source domain, parameterized by $\vct \theta \in \mathbb R^t$. To reprogram this model, we first  define a class-mapping function $h : \set{Y} \mapsto \set{\tilde Y}$ that maps each label of the target domain to a label of the source domain (\eg, the target-domain label ``0'' to the source-domain label ``tench''). We then need to optimize and embed the adversarial program in the target-domain samples. 

\begin{figure*}[t]
    \centering
    \includegraphics[width=0.99\textwidth]{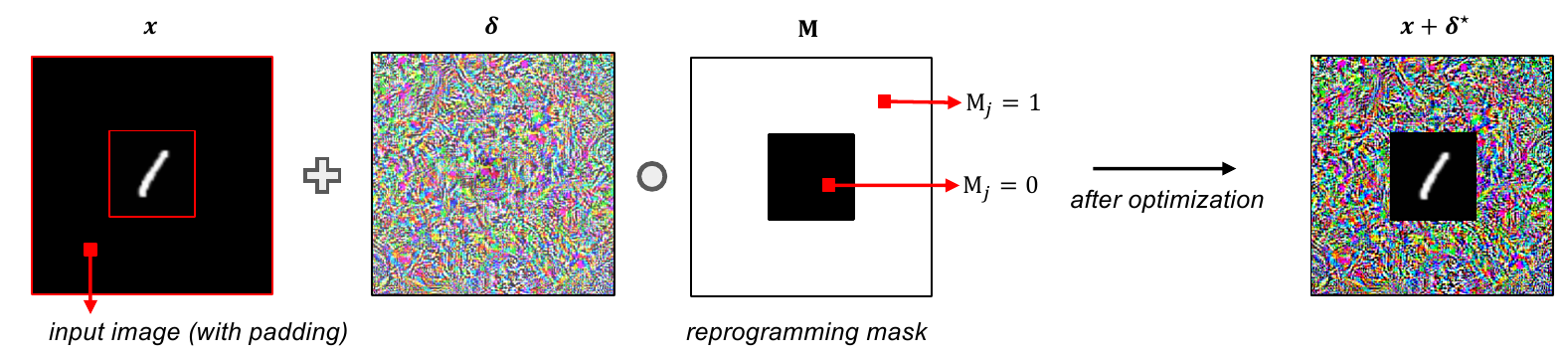}
    \caption{Reprogramming mask $\mat M$ used to restrict the adversarial perturbation to the frame surrounding the target-domain input image, initially padded with zeros.}
    \label{fig:mask}
\end{figure*}

\myparagraph{Reprogramming Mask.} Even though our model and analysis can be generalized and extended to many different kinds of adversarial programs, as discussed in Sect.~\ref{sect:extension}, we restrict our focus here on adversarial programs consisting of a frame surrounding the target-domain samples, as shown in Fig.~\ref{fig:repr-success-and-failure} and originally considered in the seminal work in~\cite{elsayed19-ICLR}. 
This means that the target-domain samples are assumed to be smaller than the source-domain samples, and padded with zeros to reach the required input size $d$; for example, MNIST handwritten digits consist of $28 \times 28 = 784$ pixels per channel, and should be padded with more than 49,000 zeros per channel to reach the input size of \imagenet models (which have $224 \times 224 = 50,176$ pixels per channel). 
We represent reprogramming masks as a binary vector $\mat M \in \{0,1\}^d$, whose values are set to 0 in the region occupied by the target-domain sample, and to 1 in the surrounding frame (\ie, the portion of the image which can be modified), as shown in Fig.~\ref{fig:mask}. 

Under these assumptions, the optimal adversarial program $\vct \delta^\star$ can be obtained by solving the following optimization problem:
\begin{equation}
\label{eq:reprogramming}
    \vct \delta^\star \in \argmin_{ \vct \delta \in [-1,1]^d} L(\vct \delta, \mathcal{T}) = \frac{1}{n} \sum_{i=1}^n \ell(\vct x_i+ \vct \delta \circ \mat M, h(y_i), \vct \theta)  \, ,
\end{equation}
where $\vct \delta$ is the adversarial program being optimized within the feasible domain $\set X=[-1,1]^d$, $\mat M \in \{0,1\}^d$ is the reprogramming mask, the $\circ$ operator denotes element-wise vector multiplication, and $\ell$ is the cross-entropy loss, which is minimized when the perturbed target-domain samples are assigned to the desired source-domain classes. Note that the constraint $\vct \delta \in [-1,1]^d$ is equivalent to upper bounding the $\ell_\infty$ norm of $\vct \delta$ as $\|\vct \delta \|_\infty \leq 1$.

\begin{algorithm}[t]
    \caption{Adversarial Reprogramming via Stochastic Gradient Descent}
    \begin{algorithmic}[1]
        \Require the target-domain dataset $\mathcal{T}=(\bm{x}_i, {y}_i)^n_{i=1}$, the model parameters $\vct \theta$, the batch size $B$, the number of iterations $N$, the step size $\eta$, and the projection operator $\Pi_{\set X}$.
        \Ensure the optimal adversarial program $\vct \delta^*$.
            \State $\vct \delta \gets \vct 0$, $\vct \delta^* \gets \vct \delta$, $\text{loss}_{ \vct \delta*} \gets \infty$\label{line:init}
            \State $t \gets 0$
            \For {$t<N$}\label{line:forloop}
                \State \text{Randomly shuffle the samples in }$\mathcal{T}$\label{line:shuffling} 
                \State $b \gets 0$
                \For{$b < \lfloor{\frac{n}{B}} \rfloor$}\label{line:forloopbatches}
                    \State $\vct g \gets \frac{1}{B} \sum_{k=B\cdot b}^{B\cdot b +B-1} \vct g_k$ \text{(average input gradient on the current batch)}\label{line:avg_grad}
                    \State $\vct \delta \gets \vct \delta -\eta \; \sign(\vct g) $ \label{line:gradient_step}
                    \State $\vct \delta \gets \Pi_{\set X}(\vct \delta)$\label{line:proj_step}
                    \State $b \gets b + 1$ 
                \EndFor
                \State $\text{loss}_{\vct \delta} = L(\vct \delta, \mathcal{T})$ \text{(compute loss as given in Eq.~\ref{eq:reprogramming})} \label{line:loss}
                    \If{$\text{loss}_{\vct \delta} < \text{loss}_{ \vct \delta*}$}\label{line:check_stop}
                    \State $\vct \delta^* \gets \vct \delta$\label{line:delta_update}
                    \State $\text{loss}_{\vct \delta*} \gets \text{loss}_{\vct \delta}$
                    \EndIf
                \State $t \gets t + 1$ 
            \EndFor
            \State {\bfseries return} $\vct \delta^\star$ \label{line:return}
    \end{algorithmic}
    \label{alg:reprogramming}
\end{algorithm}

\myparagraph{Solution Algorithm.} In this work, we use Algorithm~\ref{alg:reprogramming} to solve the optimization problem in Eq.~\eqref{eq:reprogramming} via stochastic gradient descent. This algorithm extends the Projected Gradient Descent (PGD) algorithm originally used in~\cite{madry18-iclr} to optimize adversarial perturbations within an $\epsilon$-sized $\ell_\infty$-norm constraint. 
Our algorithm iteratively updates the adversarial program $\vct \delta$ to minimize the expected loss on the target-domain samples (line~\ref{line:forloop}). In each iteration, the target-domain samples are randomly shuffled (line~\ref{line:shuffling}) and subdivided in $b$ batches of size $B$. The adversarial program is then updated by iterating over the batches (line~\ref{line:forloopbatches}). In particular, the average input gradient $\vct g$ is first computed on the batch samples (line~\ref{line:avg_grad}), and then the adversarial program is updated with an $\eta$-sized step (line~\ref{line:gradient_step}) along the sign of the negative gradient (\ie, the steepest descent direction under the given $\ell_\infty$-norm constraint). 
In our case, the gradient for the $i^{\rm th}$ sample is computed as $\vct g_i = \nabla_{\vct \delta} \ell(\vct x_i, h(y_i), \vct \theta) \circ \mat M$, \ie, including the application of the reprogramming mask $\mat M$. 
After updating $\vct \delta$, the algorithm projects the program onto the feasible space $\set X =[-1, 1]^d$, using the box-projection operator $\Pi_{\set X}$ (line~\ref{line:proj_step}). 
The algorithm finally returns the adversarial program $\vct \delta^\star$ that achieves the minimum classification loss across the whole optimization process (line~\ref{line:return}).

\subsection{A First-order Model of Adversarial Reprogramming}
\label{sec:firs-order-analysis}

We are now ready to introduce the linear model proposed in this work to better understand which underlying factors mostly affect the success of adversarial reprogramming.

\myparagraph{Linearization.} Reprogramming aims to minimize the loss in Eq.~\eqref{eq:reprogramming} to have the target-domain samples classified as desired within the source-domain classes. Let us linearize the loss in Eq.~\eqref{eq:reprogramming}: 
\begin{equation}
    \label{eq:linearized-reprogramming-loss}
     L(\vct \delta) \approx \frac{1}{n} \sum_{i=1}^n \ell(\vct x_i, h(y_i), \vct \theta) +  \frac{1}{n} \sum_{i=1}^n  \vct \delta^\T \underbrace{\nabla_{\vct x} \ell(\vct x_i, h(y_i), \vct \theta) \circ \mat M}_{\vct g_i} \, ,
\end{equation}
where $\vct g_i$ is the \textit{masked} input gradient, \ie, the input gradient computed for the $i^{\rm th}$ test sample, multiplied by the reprogramming mask $\mat M$. This amounts to zeroing the values of the input gradient for which the mask is zero. This approximation may not only hold for sufficiently-small input perturbations. It may also hold for larger perturbations if the classification function is linear or has a small curvature (e.g., if it is strongly regularized).

\myparagraph{Source and Target Domain Alignment.} The first term in Eq.~\ref{eq:linearized-reprogramming-loss}, \ie, the loss $\frac{1}{n} \sum_i \ell(\vct x_i, h(y_i), \vct \theta)$, measures how difficult is to reprogram a given machine-learning model from the source to the target domain, without applying any input perturbation. 
If this loss term is small, indeed, it means that the unperturbed samples from the target domain are already consistently assigned to the desired source-domain classes with high probability; thus, even small perturbations may provide high reprogramming accuracy. Conversely, when this one-to-one mapping between source and target classes is only weakly present, this term takes on larger values, meaning that reprogramming may be more challenging.
However, our experiments in Sect.~\ref{sec:experiments} show that this term does not play a significant role for reprogramming accuracy, since the source-domain samples and the target-domain samples are normally quite different. 

\myparagraph{Loss Decrement.} The second term in  Eq.~\eqref{eq:linearized-reprogramming-loss} is the loss decrement that can be achieved when optimizing $\vct \delta$. It can be rewritten as: 
\begin{equation}
    \Delta L(\vct \delta) =  \vct \delta^\T \frac{1}{n}\sum_i \vct g_i = \vct \delta^\T \vct g \, , 
    \label{eq:linearz}
\end{equation}
being $\vct g$ the average input gradient.
Minimizing the scalar product $\vct \delta^\T \vct g$ under the constraint $\vct \delta \in [-1,1]^d$ given in Eq.~\eqref{eq:reprogramming} is equivalent to minimizing an inner product over a unit-norm $\ell_\infty$ ball.
It is not difficult to see that, in this case, the optimal perturbation is given as $\vct \delta^\star = -{\rm sign}(\vct g)$, \ie, it is found by setting each component $\delta_j^\star$, for $j=1,\ldots,d$, equal to the negative sign of the corresponding element in $\vct g$. Substituting $\vct \delta^\star$ into Eq.~\eqref{eq:linearz}, we obtain that: 
\begin{equation}
\label{eq:reprogramming-delta-loss}
\Delta L(\vct \delta^\star) =  -{\rm sign}(\vct g)^\T \vct g = - \sum_{j=1}^d {\rm sign}(g_j)g_j =  - \sum_{j=1}^d |g_j| = -\|  \vct g \|_1 \, .
\end{equation}
This result highlights that reprogramming is more successful when the $\ell_1$-norm of the average input gradient $\vct g$ is larger, which in turn means that reprogramming is expected to work better when:
\begin{itemize}
    \item the input gradients $\vct g_i$ are more aligned, as this would increase the $\ell_1$ norm of the average input gradient $\vct g$; 
    \item the number of input dimensions $d$ is higher, as the $\ell_1$ norm of $\vct g$ scales linearly with the number of input dimensions.
\end{itemize}

\subsubsection{Gradient Alignment} 

To measure the alignment of the input gradients $\vct g_i$ w.r.t. their average $\vct g$, \mynewcomment{obtained considering the linearized loss,} we introduce the gradient-alignment metric $r$:
\begin{equation}
\label{eq:r}
    r = \frac{\| \vct g \|_1}{ \frac{1}{n} \sum_i \| \vct g_i \|_1} \in [0,1] \, .
\end{equation}
This metric equals zero only when the average vector is $\vct g = \vct 0$, while it equals one when the $\ell_1$ norm of $\vct g$ is equal to the average of the $\ell_1$ norms of the input gradients. Note that the latter case does not require the input gradients to be all equal to the average $\vct g$, conversely to what one may intuitively think. To ensure that $r=1$, indeed, it suffices that the input gradients $\vct g_i$ are \textit{orthogonal}. For example, consider a simple case in which $\vct g_1 = (0, 0, 1)$ and $\vct g_2 = (1, 0, 0)$, with $\vct g = (0.5, 0, 0.5)$. It is not difficult to see that the average $\ell_1$ norm of $\vct g_1$ and $\vct g_2$ equals 1 as well as the $\ell_1$ norm of $\vct g$.
This means that gradient alignment is maximized even if the input gradients are nearly orthogonal, which is quite likely to happen especially in high-dimensional input spaces (at least under the assumption that they are independent and randomly generated).

\subsubsection{Reprogramming Mask Size} 

The other main factor affecting the success of adversarial reprogramming, as anticipated before, is the number of input dimensions $d$. In particular, not only gradient alignment is expected to increase when the input space is high dimensional, but also the $\ell_1$ norm of the average input gradient is expected to grow linearly with the number of input dimensions $d$. Note however that, when using a reprogramming mask, the number of dimensions (\ie, pixels) used to optimize the adversarial program is not equal to $d$, but rather to the size of the reprogramming frame surrounding the input image.
We will thus control the actual size of the adversarial program in our experiments by varying the number of pixels that are used by the adversarial program, \ie, by changing the size of the reprogramming mask. The size of the reprogramming mask can be measured by simply counting the number of ones in $\mat M$, \eg, by computing $\| \mat M \|_1$.

\subsection{Extension to Other Perturbation Models} 
\label{sect:extension}

As discussed in Sect.~\ref{sec:firs-order-analysis}, the success of reprogramming mostly depends on the minimization of the second term of Eq.~\eqref{eq:linearized-reprogramming-loss}, whose minimum, under linearization, is given as $ \Delta L(\vct \delta^\star) = \vct \delta^\T \vct g = -\|  \vct g \|_1$.

Interestingly, it is not difficult to see that the $\ell_1$ norm of $\vct g$ naturally appears here as it corresponds to the dual norm of the $\ell_\infty$ norm used in Eq.~\ref{eq:reprogramming} to bound the adversarial perturbation $\vct \delta$. 
While this result may not be surprising at first sight, as the dual norm is obtained by definition as the maximization of a scalar product over a unit ball~\cite{Boyd-Vandenberghe-Convex-2004}, it allows us to extend our model to adversarial programs optimized with different perturbation constraints. For example, one may use a generic $\ell_p$-norm constraint (with $p=1,2,\infty$) bounded by a small perturbation size $\epsilon$, \ie, $\| \vct \delta\|_p \leq \epsilon$, to superimpose the adversarial program in an imperceptible manner on the input image, and found that the optimal $\Delta L(\vct \delta^\star)$ is simply given as $\Delta L(\vct \delta^\star) = -\epsilon \| \vct g \|_q$, being $q$ the dual norm of $p$.

We conclude this section by pointing out that, while similar findings have been derived in~\cite{simon19-pmlr} to study the behavior of adversarial examples in high-dimensional spaces, our model extends the analysis in~\cite{simon19-pmlr} to account for adversarial perturbations which are optimized \textit{over} different input samples (and not just on a single input image). Such perturbations do not only include adversarial programs, but they also encompass universal adversarial perturbations~\cite{moosavi17-cvpr} and robust adversarial examples~\cite{athalye18-iclr,song18-cvpr}, making our model readily applicable to study and quantify also the effectiveness of such attacks.

\section{Experimental Analysis}
\label{sec:experiments}

We report here an extensive experimental analysis aimed at evaluating the impact of the factors identified by the mathematical model presented in Sect.~\ref{sec:math-tool} on reprogramming accuracy.

\subsection{Experimental Setup}

We consider 14 different reprogramming tasks and 3 model architectures, focusing on deep neural networks trained to perform computer-vision tasks. In the following, we provide the details required to reproduce our empirical analysis.

\myparagraph{Datasets.} Our experimental analysis considers four different computer-vision datasets: two object recognition datasets, \ie, \imagenet and \cifar, and two datasets containing images of handwritten Arabic (\mnist) and Chinese (\chmnist) digits. 

\noindent \textit{\imagenet}\footnote{\url{https://www.image-net.org/}} is one of the largest publicly-available computer-vision datasets. It contains images belonging to $1,000$ categories subdivided in around 1.2 million training images and 150,000 test images. The images are collected from Internet by search engines and labeled by humans via crowdsourcing. We use this dataset only as a source-domain dataset, as many pretrained models on \imagenet are readily available.
 
\noindent \textit{\cifar}\footnote{\url{https://www.cs.toronto.edu/~kriz/cifar.html}} is a ten-class image-classification dataset made up of small resolution ($32 \times 32$) color images, subdivided into 50,000 training and 10,000 test images. 

\noindent \textit{\mnist}\footnote{\url{http://yann.lecun.com/exdb/mnist/}} is a ten-class dataset containing images of handwritten Arabic digits. It consists of grayscale images of size $28 \times 28$, subdivided in 60,000 training and 10,000 test images. 

\noindent \textit{\chmnist}\footnote{\url{http://www.pris.net.cn/introduction/teacher/lichunguang}} is a large-scale handwritten Chinese character database~\cite{zhang09-icdar}, containing grayscale images of size $64 \times 64$.
We consider only the ten classes corresponding to the Chinese digits, which result into 7,000 training and 3,000 test images.

\mynewcomment{\myparagraph{Dataset Splits.} To train the models, we have used the full training dataset of the source domain dataset. We have used 5000 samples randomly sampled from the training dataset of the target domain dataset to compute the adversarial program and 1000 samples, randomly sampled from the test set of the target domain dataset to compute the performance metrics.}

\myparagraph{Preprocessing.} We rescale the input images in~$\set X = [-1, 1]^d$, with $d = 224 \times 224 \times 3$, to match the input size of the considered models. 
This requires padding input images with zeros if they are smaller in size, and replicating the image content on each color channel for single-channel grayscale images (like in the case of MNIST and Chinese digits).

\myparagraph{Classifiers.} We consider three different neural-network architectures: the CWNet network proposed in~\cite{carlini17-sp} (Table~\ref{tab:small-dnn-architecture}); and the pretrained \mynewcomment{(and thus trained on the full training dataset)} \imagenet models AlexNet~\cite{alex12-nips} and EfficientNet~\cite{tan19-pmlr}.
All the considered models have input size $d = 224 \times 224 \times 3$, and apply z-score batch normalization before processing the input samples.
CWNet is only trained using MNIST and HCL2000 as the source-domain data.
AlexNet and EfficientNet are instead used with \imagenet, MNIST and HCL2000 as source-domain datasets. 
For AlexNet and EfficientNet, we also consider a setting in which their weights are set to random values, which amounts to considering their \textit{untrained} versions as done in~\cite{elsayed19-ICLR}.
When training the considered models on MNIST and HCL2000, we run stochastic gradient descent for $10$ epochs, with step size, momentum, and batch size respectively equal to $0.001$, $0.9$, and $10$. 
For the step size, the number of epochs, and batch size, we chose the values that lead to a higher accuracy on a validation dataset of 1000 samples sampled from the training dataset belonging to the source domain.

\begin{table}[tb]
\centering
\caption{Architecture of the CWNet network trained on digit images in~\cite{carlini17-sp}.}
\label{tab:small-dnn-architecture}
%\resizebox{0.45\textwidth}{!}{%
\begin{tabular}{l|c}
    \toprule
    Layer Type & Dimension\\
    \midrule
    Conv. + ReLU & 32 filters ($3\times3$) \\
    Conv. + ReLU & 32 filters ($3\times3$) \\
    Max Pooling & $2\times2$ \\
    Conv. + ReLU & 64 filters ($3\times3$) \\
    Conv. + Dropout (0.5\%) + ReLU & 64 filters ($3\times3$) \\
    Max Pooling & $2\times2$ \\
    Fully Connected + ReLU & 200 units \\
    Fully Connected + ReLU & 200 units \\
    Softmax  & 10 units \\
    \bottomrule
\end{tabular}
%}
\end{table}

\myparagraph{Adversarial Programs.} 
To optimize the adversarial program $\vct \delta$, we use Algorithm~\ref{alg:reprogramming}. Before optimizing it, we fix the class-mapping $h$ as a function that maps orderly the first ten classes of the source dataset to the first ten classes of the target dataset, as done in~\cite{elsayed19-ICLR}. (The first class of the source domain dataset is mapped to the first class of the target domain dataset, the second of the source domain in the second of the target domain dataset, etc.). As in~\cite{elsayed19-ICLR}, for the datasets having more than classes than the target domain dataset, which in our case is only \imagenet, which contains more than ten classes, we only consider the first ten classes (\ie, tench, goldfish, white shark, tiger shark, hammerhead, electric ray, stingray, cock, hen, ostrich). Namely, we crop the output layer so that it contains 10 elements (one for each of the abovementioned ten classes, and we consider as the predicted class, the one that received a higher score between these ten classes). Note that the choice of the mapping function may affect the accuracy; however as we will explain in the following, it will not impact our claims.
We set the step size $\eta = 0.005$, $N=100$ epochs, and we use batches of $B=50$ samples, sampled from a larger set of $5,000$ images randomly drawn from the training set of the target-domain dataset $\set T$.
%Moreover, as the adversarial program is expected to generalize to unseen samples, we do not compute the loss in Eq.~\eqref{eq:reprogramming} with the same samples used to optimize the perturbation, but rather use a separate set of $5,000$ images sampled again from the target-domain training set. This should provide a better, unbiased evaluation of the effectiveness of the adversarial program.

\myparagraph{Performance Metrics.} We consider different metrics to evaluate the performance of reprogramming along with the underlying factors affecting its success. In particular, we consider three metrics: 
\begin{itemize}
    \item \textit{reprogramming accuracy (RA)}, \ie the ratio of samples of the target-domain dataset correctly classified after the application of the adversarial program. A higher RA means that the adversarial program is more effective in repurposing the model, thus the higher, the better;
    \item \textit{domain alignment (DA)}, evaluated as the model's accuracy on the target-domain samples padded with zeros, \ie, before optimizing the adversarial program $\vct \delta$. A higher DA means that the first term of Eq.~\ref{eq:linearized-reprogramming-loss} is low, which means that the unperturbed samples from the target domain are already consistently assigned to the desired source-domain classes with high probability. Thus, even small perturbations may provide high reprogramming accuracy.
    \item \textit{gradient alignment (Eq.~\ref{eq:r}) before ($r_0$) and after ($r_N$) reprogramming}, \ie, after, respectively, the first and the last iteration of Algorithm~\ref{alg:reprogramming}. Accordingly to our analysis, the gradient alignment helps perform reprogramming; thus, it is better when it is high. (Intuitively, if the perturbations that should be applied to each sample to reprogram the model are aligned, finding a single universal perturbation to reprogram the model is easier).
\end{itemize}

\subsection{Experimental Results}

\begin{table*}[t]
\centering
\caption{Results of reprogramming AlexNet, EfficientNet, and CWNet from different source ($\set S$) to target ($\set T$) domains. U means that the network has not been trained (its weights are randomly initialized). For each reprogramming task, the table reports domain alignment (DA), reprogramming accuracy (RA), and gradient alignment (Eq.~\ref{eq:r}) computed before ($r_0$) and after ($r_N$) optimizing the adversarial program $\vct \delta$. For all the reported metrics, the higher the value the better. The cases in which reprogramming works well/poorly/badly are highlighted in green/yellow/red.}
\label{tab:results-mask-size-224}
%\resizebox{\textwidth}{!}{%
\begin{tabular}{ccc|c|c|c|c}
\hline
$\set S$ & $\set T$ & Model        & ${\rm DA}$ & ${\rm RA}$ & $r_0$ & $r_N$ \\ \hline
\rowcolor[HTML]{D9EFD9}\imagenet & \chmnist & EfficientNet & $5.3\%$                   & $98.1\%$                 & $18.4\%$  & $21.14\%$ \\
\rowcolor[HTML]{D9EFD9}\imagenet & \chmnist & AlexNet      & $2.5\%$                   & $97.2\%$                 & $19.3\%$  & $19.35\%$ \\
\rowcolor[HTML]{D9EFD9}\imagenet & \mnist   & EfficientNet & $14.3\%$                  & $90.6\%$                 & $17.5\%$  & $20.33\%$ \\
\rowcolor[HTML]{D9EFD9}\imagenet & \mnist   & AlexNet      & $11.6\%$                  & $90.1\%$                 & $29.0\%$  & $18.75\%$ \\ \hline
\rowcolor[HTML]{EFEFCD}\imagenet & \cifar   & EfficientNet & $10.2\%$                  & $51.1\%$                 & $13.5\%$  & $10.54\%$ \\
\rowcolor[HTML]{EFEFCD}\imagenet & \cifar   & AlexNet      & $9.3\%$                   & $46.0\%$                 & $13.6\%$  & $11.28\%$ \\
\rowcolor[HTML]{EFEFCD}\mnist    & \chmnist & AlexNet      & $12.8\%$                  & $49.1\%$                 & $14.5\%$  & $9.11\%$  \\
\rowcolor[HTML]{EFEFCD}\mnist    & \chmnist & CWNet          & $9.5\%$                   & $45.3\%$                 & $16.4\%$  & $13.59\%$ \\ \hline
\rowcolor[HTML]{FFCCCC}\chmnist  & \mnist   & AlexNet      & $9.9\%$                   & $21.8\%$                 & $92.1\%$  & $5.27\%$  \\
\rowcolor[HTML]{FFCCCC}\chmnist  & \mnist   & CWNet          & $9.9\%$                   & $19.1\%$                 & $92.8\%$  & $4.82\%$  \\
\rowcolor[HTML]{FFCCCC}U         & \mnist   & AlexNet      & $11.8\%$                  & $20.5\%$                 & $5.0\%$   & $6.0\%$   \\
\rowcolor[HTML]{FFCCCC}U         & \chmnist & AlexNet      & $9.0\%$                   & $28.7\%$                 & $6.7\%$   & $6.36\%$  \\
\rowcolor[HTML]{FFCCCC}U         & \mnist   & EfficientNet & $11.8\%$                  & $11.8\%$                 & $4.0\%$   & $3.38\%$  \\
\rowcolor[HTML]{FFCCCC}U         & \chmnist  & EfficientNet & $9.0\%$                   & $9.0\%$                  & $6.9\%$   & $5.72\%$  \\ 
\hline
\end{tabular}
%}
\end{table*}

\begin{figure*}[t]
	\centering
	%quando passiamo a due colonne commentare riga con resizebox 
	\resizebox{0.99\columnwidth}{!}
	{\begin{tabular}{ccccc}
			\centering
			EfficientNet 
			& AlexNet 
			& EfficientNet 
			& AlexNet 
			& EfficientNet 
			\\ (\imagenet$\rightarrow$\chmnist)
			& (\imagenet$\rightarrow$\chmnist)
			& (\imagenet$\rightarrow$\mnist)
			& (\imagenet$\rightarrow$\mnist)
			& (\imagenet$\rightarrow$\cifar)
			\\		
			% quando passiamo a 2 colonne 15 come dimensione
			\includegraphics[width=0.20\textwidth]{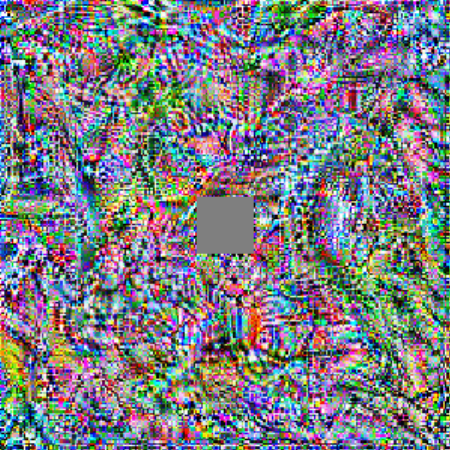}	&
        	\includegraphics[width=0.20\textwidth]{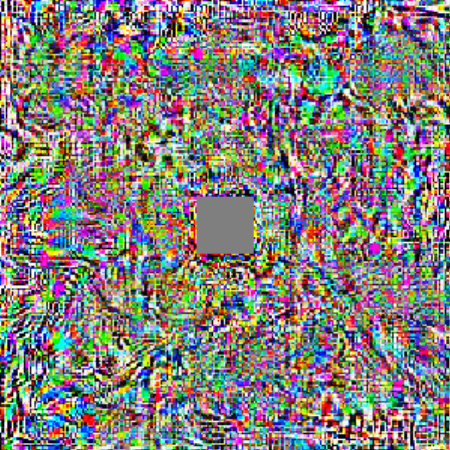} &
        	\includegraphics[width=0.20\textwidth]{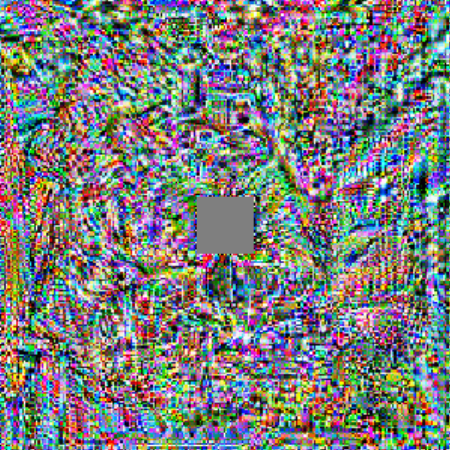} &
        	\includegraphics[width=0.20\textwidth]{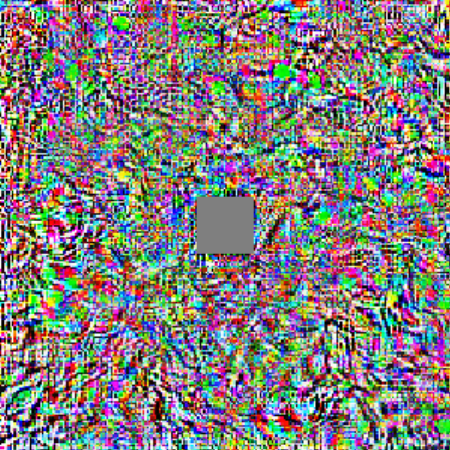} &        	
        	\includegraphics[width=0.20\textwidth]{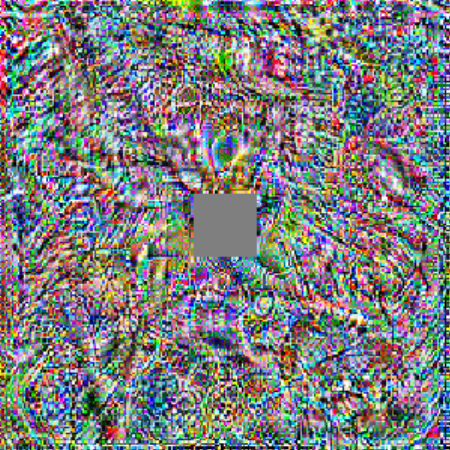}  \\		
			AlexNet 
			& AlexNet 
			& CWNet
			& AlexNet 
			& CWNet 
			\\
			(\imagenet$\rightarrow$\cifar)
			& (\mnist$\rightarrow$\chmnist)
			& (\mnist$\rightarrow$\chmnist)
			& (\chmnist$\rightarrow$\mnist)
			& (\chmnist$\rightarrow$\mnist)
			\\			
			\includegraphics[width=0.20\textwidth]{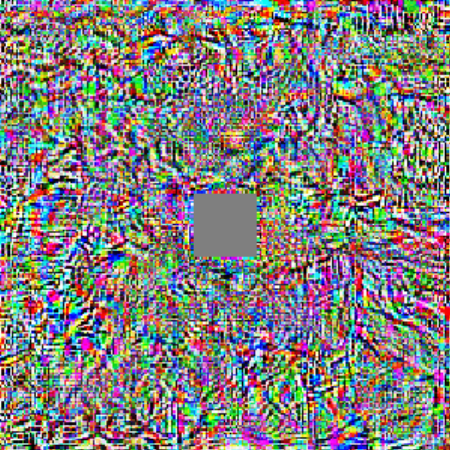} &
			\includegraphics[width=0.20\textwidth]{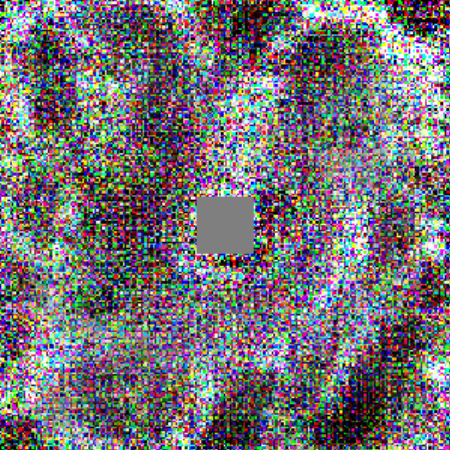} &
        	\includegraphics[width=0.20\textwidth]{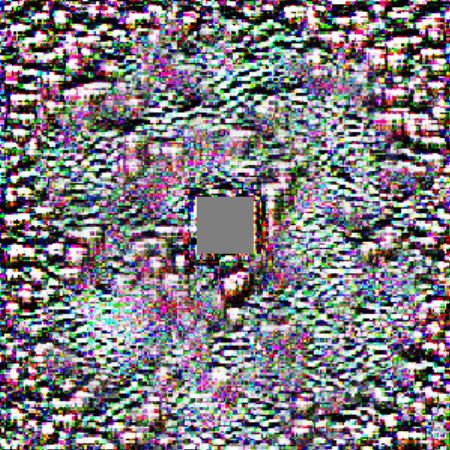} &
        	\includegraphics[width=0.20\textwidth]{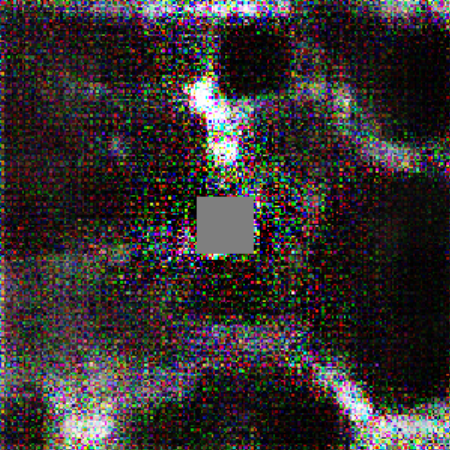} &
        	\includegraphics[width=0.20\textwidth]{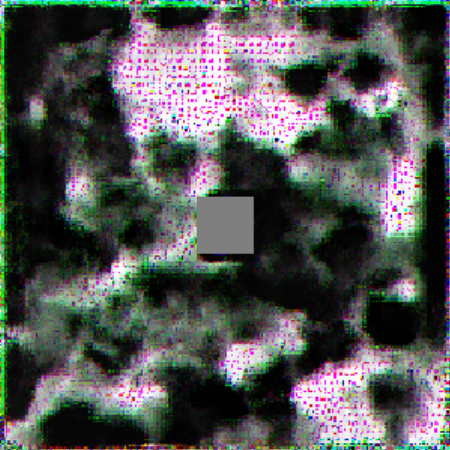}
		\end{tabular}}
		
		% rimuovere quando passo a due colonne
		\resizebox{0.58\columnwidth}{!}
		{\begin{tabular}{cccc}
	\centering
	AlexNet &
	AlexNet &
	EfficientNet &
	EfficientNet 
	\\	
	(U $\rightarrow$ \mnist) &
	(U $\rightarrow$ \chmnist) &
	(U $\rightarrow$ \mnist) &
	(U $\rightarrow$ \chmnist)
	\\		
	%quando passo a due colonne riportare a 15
	\includegraphics[width=0.20\textwidth]{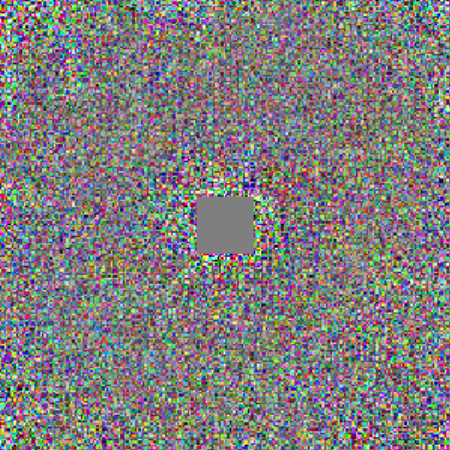} &
	\includegraphics[width=0.20\textwidth]{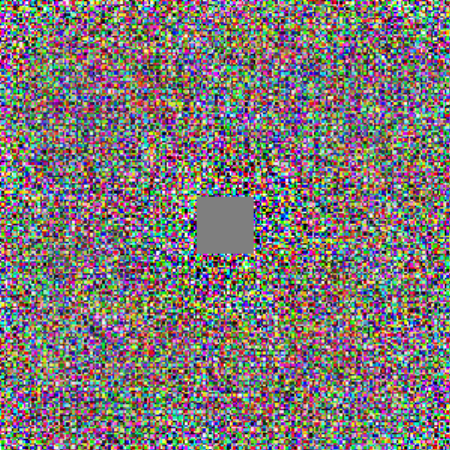}&
	\includegraphics[width=0.20\textwidth]{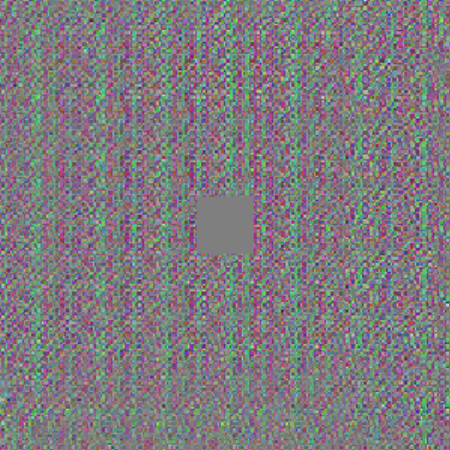}	&
	\includegraphics[width=0.20\textwidth]{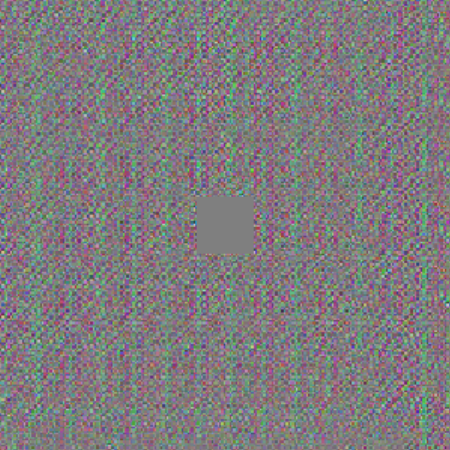}		
		\end{tabular}}
    \caption{Adversarial programs optimized to repurpose the networks in Table~\ref{tab:results-mask-size-224} from different source to target ($\set S \rightarrow \set T$) domains.}
		\label{fig:programs}
	\end{figure*}

The experimental results on the given 14 reprogramming tasks are reported in Table~\ref{tab:results-mask-size-224}, while the corresponding adversarial programs are shown in Fig.~\ref{fig:programs}.
As highlighted in Table~\ref{tab:results-mask-size-224} with different colors, adversarial reprogramming may exhibit different \textit{reprogramming accuracy} (RA) values: it may work remarkably well (${\rm RA} \geq 90\%$), it may work poorly (${\rm RA} \approx 50\%$), or it may even completely fail (${\rm RA} \leq 30\%$). 
In the remainder of this section, we analyze the impact on reprogramming accuracy of the main factors identified by our mathematical model in Sect.~\ref{sec:math-tool}, \ie, ($i$) the alignment between source and target domain, ($ii$) the alignment of the input gradients, and ($iii$) the size of the reprogramming mask. In the following paragraphs (one for each of these factors), we recap our claims (in italic), then present the empirical results. Note that from our experimental analysis, all these factors help perform reprogramming. However, in the following, we assess to which extent they contribute to the reprogramming success in practice.

%%%%%%%%%%%%%%%%%%%%
%%%%%%%%%%%%%%%%

\begin{figure*}[t]
	\centering
    \resizebox{0.99\columnwidth}{!}
	{\begin{tabular}{cccc}
	\multicolumn{2}{c}{\Large AlexNet (\imagenet$\rightarrow$\chmnist),  ${\rm RA}$=97.2\%} & \multicolumn{2}{c}{\Large AlexNet (MNIST$\rightarrow$\chmnist), ${\rm RA}$=49.1\%} \\
	\includegraphics[trim=0 0 0 0, clip, width=0.51\textwidth]{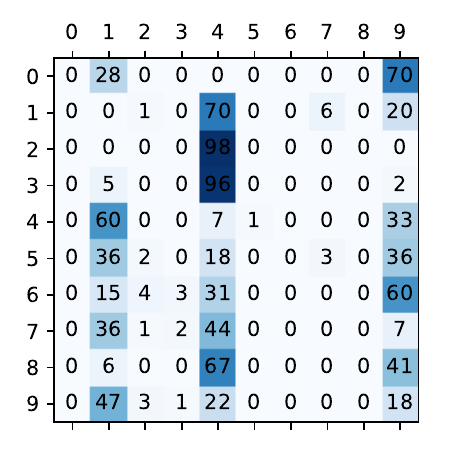} &
	\includegraphics[trim=25 0 0 0, clip, width=0.45\textwidth]{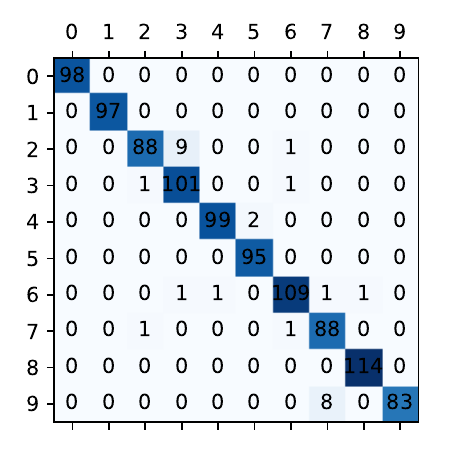} &
	
	\includegraphics[trim=25 0 0 0, clip, width=0.45\textwidth]{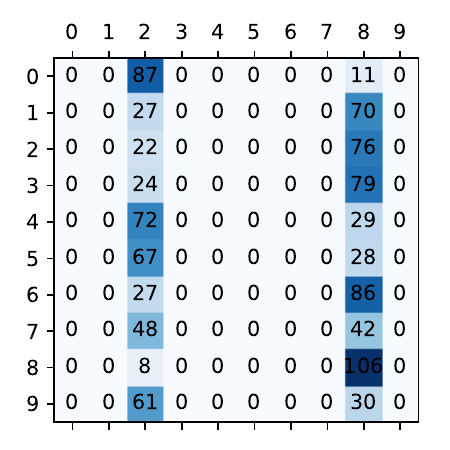} &
	\includegraphics[trim=25 0 0 0, clip, width=0.45\textwidth]{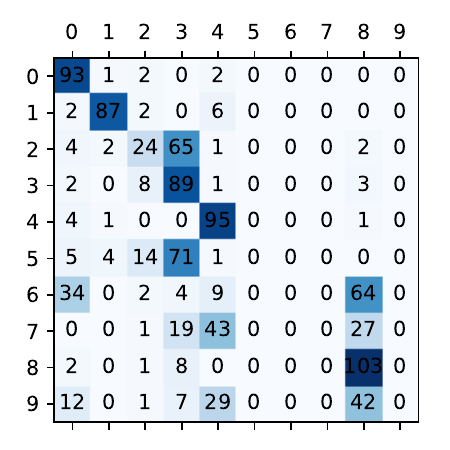} \\	
	
	\multicolumn{2}{c}{\Large AlexNet (HCL2000$\rightarrow$\mnist), ${\rm RA}$=21.8\%} & \multicolumn{2}{c}{\Large AlexNet (U$\rightarrow$\mnist), ${\rm RA}$=20.5\%}
	\\
	\includegraphics[trim=0 0 0 0, clip, width=0.51\textwidth]{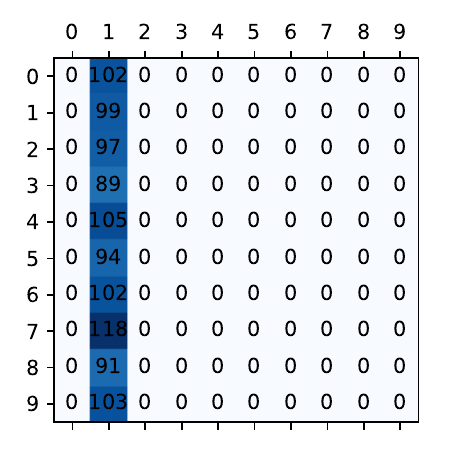} &
	\includegraphics[trim=25 0 0 0, clip, width=0.45\textwidth]{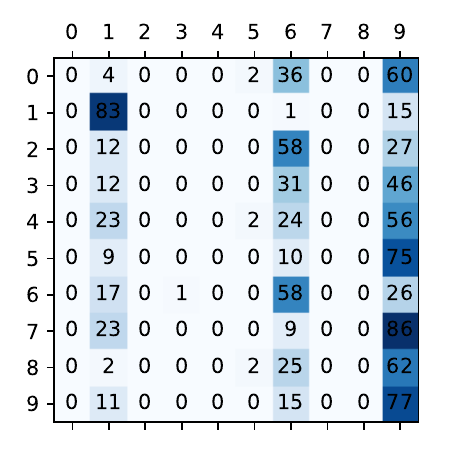} &
	
	\includegraphics[trim=25 0 0 0, clip, width=0.45\textwidth]{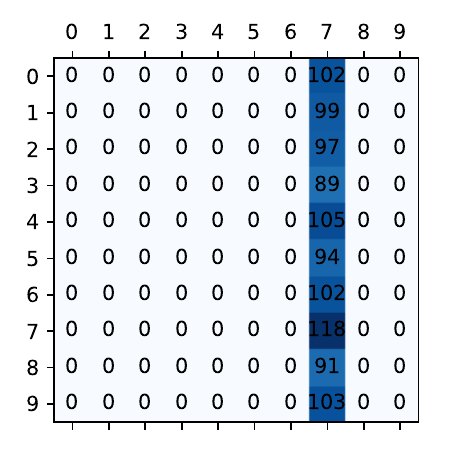} &
	\includegraphics[trim=25 0 0 0, clip, width=0.45\textwidth]{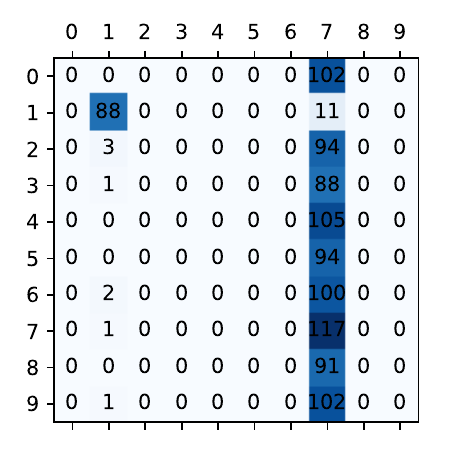}	\\	
	\end{tabular}}
	
	\caption{Confusion matrices (true-vs-predicted classes) on four representative cases. For each case, we report the confusion matrix before (\textit{left}) and after (\textit{right}) reprogramming.}
	\label{fig:confusion_matrix}
\end{figure*}

%\noindent \textit
\myparagraph{Source and Target Domain Alignment.} Let us now analyze the influence of the source-target domain alignment on reprogramming accuracy. \emph{As we have explained in Sect.~\ref{sec:math-tool}, reprogramming accuracy also depends on the classifier loss on the target-domain dataset before reprogramming (the first term of Eq.~\ref{eq:linearized-reprogramming-loss}). (It determines the reprogramming accuracy before optimizing the program).} In Table~\ref{tab:results-mask-size-224}, we report the accuracy of the classifier on the target-domain dataset before reprogramming, referred to as \textit{domain alignment} (${\rm DA}$). The table shows that DA is usually quite low (around 10\%), even when reprogramming accuracy is very high (${\rm RA}> 90\%$), which means that the source and target alignment is not correlated with the reprogramming accuracy. Namely, this factor does not have a prominent impact on reprogramming accuracy. Even if the source and target domain are quite different, reprogramming can be successful because the program optimization impacts more than the reprogramming accuracy of the model on the target domain before optimizing the program.
To explain why domain alignment is low, in Fig.~\ref{fig:confusion_matrix} we report the confusion matrices computed before (left plot) and after (right plot) reprogramming. These matrices show that, before reprogramming, the target-domain samples are often assigned to a single class or a few classes. The reason is that, before reprogramming, the target-domain samples are simply padded with zeros, thus being more likely to be classified similarly.

\myparagraph{Gradient Alignment.} Here, we analyze the impact of gradient alignment on reprogramming accuracy. \emph{As we have explained in Sect.~\ref{sec:math-tool}, the gradient alignment influences how much the reprogramming accuracy increase when optimizing the program (the higher, the better)}.
For each classifier and reprogramming task, we compute the gradient-alignment metric $r$ (Eq.~\ref{eq:r}) before ($r_0$) and after reprogramming ($r_N$). 
The reason is that, as previously explained, before reprogramming, the target-domain samples are simply padded with zeros and tend to be assigned to one or few classes. Accordingly, the input gradients computed before optimizing the program are not expected to be really informative, while they are expected to be more informative when the program is optimized and reprogramming accuracy starts increasing.

\begin{figure}[t]	
	\centering
	\includegraphics[trim=0 0 0 0, clip, width=.36\textwidth]{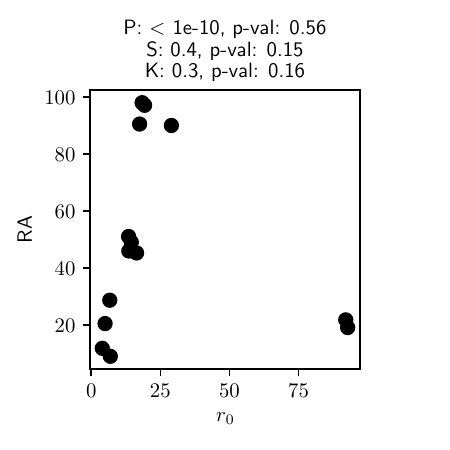}
	\includegraphics[trim=0 0 0 0, clip, width=.36\textwidth]{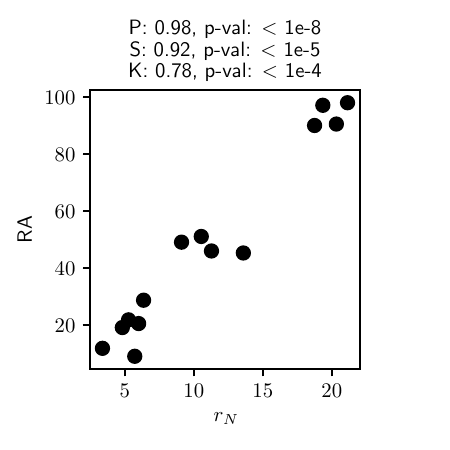}
	\caption{Correlation between the reprogramming accuracy (${\rm RA}$) vs the gradient alignment computed before $r_0$ (left) and after $r_N$ (right) optimizing the program.}
	\label{fig:scatter}
\end{figure}

The values of $r_0$ and $r_N$ for the given reprogramming tasks are reported in Table~\ref{tab:results-mask-size-224}. In Fig.~\ref{fig:scatter} we also report the correlation between  gradient alignment and reprogramming accuracy, computed using Pearson (P), Spearman (S), and Kendall (K) methods, along with the corresponding permutation tests and $p$-values. The correlation between reprogramming accuracy ${\rm RA}$ and gradient alignment before reprogramming ($r_0$) is not significant ($p>0.05$) mostly due to the presence of two outlying observations (\ie, AlexNet \chmnist $\rightarrow$ MNIST, and CWNet \chmnist $\rightarrow$ MNIST). The correlation values are much higher and statistically significant, instead, when considering gradient alignment after reprogramming ($r_N$), \eg, the correlation computed with the Pearson coefficient is 0.98 with $p < 1e-8$. These results show that gradient alignment, especially when computed after reprogramming ($r_N$), is strongly and positively correlated with reprogramming accuracy, thus confirming the intuition provided by our mathematical model.

%trim = left bottom right top

\begin{figure}[t]	
	\centering
	\includegraphics[trim=0 0 0 0, clip, width=.38\textwidth]{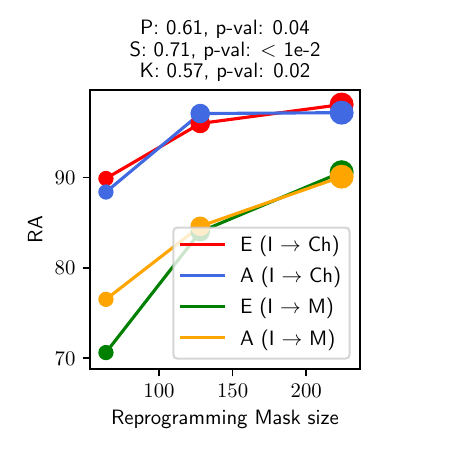}
	\includegraphics[trim=0 0 0 0, clip, width=.38\textwidth]{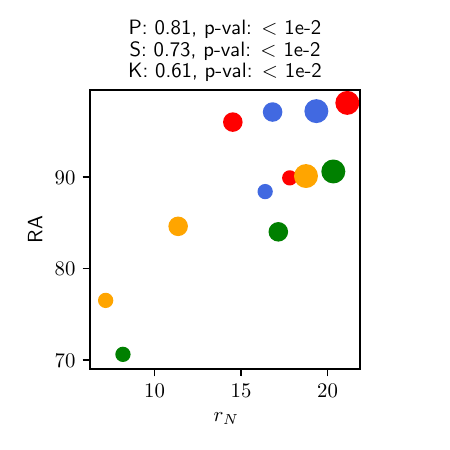}
	
	\caption{Reprogramming accuracy (${\rm RA}$) versus reprogramming mask size (left) and the gradient alignment computed after reprogramming $r_N$ (right). Considering  AlexNet (A), EfficientNet (E), and the datasets: \imagenet (I), \chmnist (Ch), \mnist (M).}
	\label{fig:scatters-different-masks}
\end{figure}

\myparagraph{Reprogramming Mask Size.} We finally analyze the impact of the reprogramming mask size on reprogramming accuracy. \emph{As we have explained in Sect.~\ref{sec:math-tool}, the reprogramming mask size influences how much the reprogramming accuracy increase when optimizing the program (the higher, the better)}.
%The classifier that we consider receives inputs with the dimensions of $224 \times 224 \times 3$. As explained in Sect.~\ref{sec:math-tool}, the reprogramming mask size that we consider in this work (if not differently specified) is equal to the input dimension.   
To this end, we consider reprogramming masks of increasing sizes, using 64, 128, and 224 as their width and height values.
We report reprogramming accuracy and the proposed measures computed for each classifier and reprogramming task in Table~\ref{tab:results-different-mask-sizes}, and the corresponding correlation tests in Fig.~\ref{fig:scatters-different-masks}. 
While the right plot in Fig.~\ref{fig:scatters-different-masks} shows again that reprogramming accuracy is strongly correlated with the gradient alignment $r_N$, also for such reprogramming cases, the left plot shows that reprogramming accuracy is also correlated with the reprogramming mask size, confirming again the soundness of the proposed mathematical model. 
From the values in Table~\ref{tab:results-different-mask-sizes}, it should also be noted that the reprogramming mask size can have a relevant impact on reprogramming accuracy, \eg, in the case AlexNet \imagenet $\rightarrow$ \mnist, there is an accuracy difference of $15\%$ between the performance obtained with the smallest and that obtained with the largest mask size considered.

\begin{table*}[t]
\centering
\caption{Results for programs with different reprogramming mask sizes. See the caption of Fig.~\ref{tab:results-mask-size-224} for further
details.}  % TODO: use (\%)
\label{tab:results-different-mask-sizes}
\begin{tabular}{cccc|c|c|c}
\toprule
Mask Size & $\set S$  & $\set T$ & Model        & RA & $r_0$ & $r_N$ \\ \hline
224           & \imagenet & \chmnist & EfficientNet & 98.1\%                  & 18.4\%    & 21.14\%   \\
128           & \imagenet & \chmnist & EfficientNet & 96.0\%                  & 18.5\%    & 14.52\%   \\
64            & \imagenet & \chmnist & EfficientNet & 89.9\%                  & 18.6\%    & 17.81\%   \\ \hline
224           & \imagenet & \chmnist & AlexNet      & 97.2\%                  & 19.3\%    & 19.35\%   \\
128           & \imagenet & \chmnist & AlexNet      & 97.1\%                  & 17.4\%    & 16.82\%   \\
64            & \imagenet & \chmnist & AlexNet      & 88.4\%                  & 16.1\%    & 16.39\%   \\ \hline
224           & \imagenet & \mnist   & EfficientNet & 90.6\%                  & 17.5\%    & 20.33\%   \\
128           & \imagenet & \mnist   & EfficientNet & 84.0\%                  & 18.2\%    & 17.15\%   \\
64            & \imagenet & \mnist   & EfficientNet & 70.6\%                  & 17.3\%    & 8.17\%    \\ \hline
224           & \imagenet & \mnist   & AlexNet      & 90.1\%                  & 29.0\%    & 18.75\%   \\
128           & \imagenet & \mnist   & AlexNet      & 84.6\%                  & 17.0\%    & 11.36\%   \\
64            & \imagenet & \mnist   & AlexNet      & 76.5\%                  & 16.7\%    & 7.17\%    \\
\bottomrule
\end{tabular}
%}
\end{table*}
%$0.38\pm0.006$

%trim = left bottom right top

\myparagraph{On the Choice of the Label Mapping.} The label mapping function can affect accuracy, as shown in~\cite{tsai20-pmlr}. As explained by the authors of that paper, the accuracy of the reprogrammed model on the target task can be improved in two ways. The first is to map the labels depending on the models' predictions before reprogramming (see ``frequency-based mapping'' in~\cite{tsai20-pmlr}), and this is always applicable. The second is the so-called multi-label mapping that maps one single label of the target domain into multiple labels of the source domain. This method is applicable only when the number of classes of the source domain is bigger than the number of classes of the target domain.
The only reprogramming tasks (between the ones considered in our work) to which this technique is applicable are the ones that reprogram models trained on \imagenet to work on the \chmnist, \mnist, and \cifar datasets. These are cases in which reprogramming works better; hence, an eventual accuracy improvement would not affect our study's outcome.
In the other reprogramming tasks, in which reprogramming does not work, the source datasets have the same number of classes as the target datasets; therefore, the only technique that could be applied is frequency-based mapping. However, this technique will only have a minimal impact on the accuracy because, as visible from the confusion matrices in Figure~\ref{fig:confusion_matrix}, the model assigns all the samples of the target domain to very few classes. We have checked, and it is also true for the confusion matrices we have not reported in the manuscript. Therefore, frequency-based mapping can find the best mapping so that the majority of the samples belonging to those classes are predicted correctly. However, this will help with at most 3 of the ten classes (the other classes are predicted only sporadically). Moreover, from [4], it is clear that this technique can achieve only marginal improvements. Using this technique instead of random mapping, they have always gained less than 5\% accuracy (see the difference between ``AR + Rand mapping'' and ``AR + Freq. Mapping'' in Figure 3 of~\cite{tsai20-pmlr}. Therefore, also, in this case, applying this technique, our claim will not change.

\myparagraph{Summary of the Results.} In this work, we develop a first-order linear model that expresses the reprogramming optimization problem as a sum of two terms. The first term measures how difficult it is to reprogram a given machine-learning model from the source to the target domain without optimizing the adversarial program. The second term is the loss decrement that we can obtain by optimizing the adversarial program and, as we showed, depends on the size of the average input gradient. From this mathematical model and our empirical study of the relationship between these terms and the reprogramming accuracy, we derived the answer to the question asked in the title: why adversarial reprogramming works, when it fails, and how to tell the difference?
Adversarial reprogramming consists of applying a single (universal) perturbation to the target-domain samples to move them into a region of the decision space where they are classified as desired.
It works because, when the size of the average input gradient of the target model is large enough, a small universal perturbation is enough to reprogram the model. Which happens when (i) the gradients that should be applied to every single sample to reprogram the model are sufficiently aligned, and (ii) the size of the reprogramming maks is sufficiently large. The latter plays a much more marginal role than the former, as small mask sizes are usually already sufficient to achieve good reprogramming performance. Also, the source and target domains alignment does not have a relevant impact on the reprogramming accuracy, as it works successfully even when the source and target domains are poorly aligned. 
Adversarial reprogramming fails when the size of the average input gradient is not large enough, as in the paradigmatic example of reprogramming failure we described in the introduction (Fig. 1 bottom). In that example, the goal was to reprogram a small convolutional neural network (CWNet), trained to recognize handwritten Chinese digits ( depicted in Fig. 1 (bottom)), to recognize the MNIST handwritten (Arabic) digits. However, in that case, reprogramming fails as the gradient alignment after reprogramming ($r_N$), which we have shown to be correlated with the reprogramming accuracy, is very low (see Table~\ref{tab:results-mask-size-224}). The intuition is that if the perturbations you should apply to every sample to reprogram the model are not sufficiently similar (the input gradients are not aligned), you cannot find a single perturbation able to reprogram the model. Namely, a single perturbation that can be employed to move the samples of the different classes into a region of the decision space where they are classified as desired. 
To summarize, the main findings of our extensive experimental analysis are:
\begin{itemize}
\item source and target domain alignment, in practice, does not affect reprogramming accuracy;
\item gradient alignment plays a key role in the success of reprogramming; 
\item larger reprogramming mask sizes facilitate reprogramming but have a much more marginal impact on its success.
\end{itemize}

\myparagraph{Additional Comments.} Here, we provide further comments on the experimental results.
On the basis of the reprogramming accuracy reported in Table~\ref{tab:results-mask-size-224}, we argue that reprogramming works better when the reprogramming task is simple and the function learned by the target classifier is complex. 
Table~\ref{tab:results-mask-size-224} also shows that reprogramming works well when the target model is a large model trained on a large and complex dataset and reprogrammed to perform a simple task (such as handwritten digit classification). On the other hand, it does not work well: (i) when learning on the target domain represents a complex task, (ii) when the target model is small, and (iii) when the target model is complex but trained on a simple and small dataset.
Whereas the first and the second case support our hypothesis, the latter deserves further clarification. When a complex classifier is trained on a small and simple dataset, the representations learned by the different network layers tend to become very similar~\cite{nguyen21-iclr}, and some of the layers thus do not alter the classifier prediction~\cite{srivastava15-nips}. Therefore, although the classifier is complex, the function learned by the classifier remains simple.

\section{Related Work}\label{sec:related}

In this section, we briefly review related work on reprogramming. We then focus on attacks that optimize adversarial perturbations, including universal adversarial perturbations and robust physical-world adversarial examples. Finally, we review work that provides additional insights into the vulnerability of machine-learning models to adversarial perturbations based on different first-order linear analyses.

\subsection{Adversarial Reprogramming} 
Adversarial reprogramming has been originally proposed in~\cite{elsayed19-ICLR}.
The authors have empirically assessed the performance of adversarial reprogramming using different trained and untrained deep neural networks. They showed that reprogramming usually fails when applied to untrained networks (\ie, neural networks with random weights), whereas it works when the target model is trained. In the latter case, reprogramming works even when the attacker can manipulate only a small subset of the image pixels. However, the authors have not explained why reprogramming works and when it fails, leaving this analysis to future work. Although subsequent work has successfully applied reprogramming in different scenarios~\cite{tsai20-pmlr,neekhara19-EMNLP,neekhara21-arxiv}, no work has analyzed the reasons behind its success and failure. It is also worth remarking that, \mynewcomment{while Yang et al.~\cite{yang2021voice2series} have identified some factors affecting why reprogramming works in the audio domain, they have not discussed when it can fail and which factors may lead to its failure. We do believe that our work is thus the first to provide a more detailed and quantitative analysis of the impact of the main factors underlying the success and failure of adversarial reprogramming.} 
%However, in our work, we also propose a metric to specifically measure the impact of each of the two terms, and support our assumptions with strong experimental evidence. To the best of our knowledge, our work is thus the first to provide a quantitatively analyze the impact of the main factors underlying the success of adversarial reprogramming. Contrarily to what was claimed in Yang et al.~\cite{yang2021voice2series}, it shows that the diversity between the two domains does not play a key role, at least when considering the same task, which in our works is classification. 

\subsection{Universal and Robust Adversarial Perturbations}

Gradient-based attacks on machine-learning models~\cite{biggio12-icml,biggio13-ecml,biggio18,joseph18-advml-book} have been demonstrated in a variety of application domains, including computer vision and security-related tasks~\cite{melis17-vipar,demontis19-tdsc,pierazzi20-sp,maiorca20-cs,anderson18,demetrio20-tifs}, even before the independent discovery of adversarial examples against deep neural networks~\cite{szegedy14-iclr}. 

While earlier work has focused on optimizing a different adversarial perturbation for each input sample, in~\cite{moosavi17-cvpr} the authors have shown that it is even possible to optimize a single, \textit{universal adversarial perturbation}, \ie, a fixed perturbation that can be applied to many different input samples to have them misclassified as desired. The underlying idea is to optimize the perturbation on different input samples, similar to the idea behind the optimization of robust physical-world adversarial examples~\cite{athalye18-iclr,song18-cvpr}, \ie, to optimize the adversarial perturbation over different transformations of the input image (\eg, subject to changes in pose, rotation, illumination).

In this work, we argue that the mathematical formulation of universal perturbations, robust adversarial examples, and adversarial reprogramming is essentially the same, \ie, all these attacks require optimizing the adversarial perturbation by averaging the loss function over different input images (even though reprogramming optimizes the perturbation to repurpose a model, while the other attacks aim to have input samples misclassified).
For this reason, we believe that our analysis can be readily applied in future work to provide a better understanding of the effectiveness of both universal adversarial perturbations and robust adversarial examples.

\subsection{First-order Analysis of Adversarial Perturbations}
\label{sect:rel-first-order}

Previous work has analyzed the vulnerability of neural networks against adversarial examples and universal adversarial perturbations. The authors of~\cite{moosavi-dezfooli18-iclr} have proven the existence of small universal perturbations, attributing them to the low curvature of the decision boundaries of deep neural networks~\cite{Fawzi18-cvpr, Moosavi-Dezfooli19-CVPR}.
The work in~\cite{simon19-pmlr} is probably the closest one to ours, as it also builds on the previously-proposed idea of modeling the optimization of adversarial examples as a linear problem~\cite{goodfellow15-iclr}.
The same idea has also been explored to develop robust methods based on regularizing the input gradients~\cite{lyu15-icdm,varga17-arxiv,ross18}, as well as to provide deeper insights on why adversarial examples can often \textit{transfer} across different models~\cite{demontis19-usenix}.
The main difference between our work and the work in~\cite{simon19-pmlr} is that our model extends their analysis to also encompass adversarial perturbations which are optimized on different samples, thereby not only including adversarial examples, but also adversarial reprogramming, universal adversarial perturbations, and robust physical-world adversarial examples.

%Work on reprogramming (Goodfellow) \cite{elsayed19-ICLR}, and connections to other work on attacks (e.g., universal perturbations \cite{moosavi17-cvpr}) / parasite computing.

%Other work on similar analysis (first-order vulnerability of NNs \cite{simon19-pmlr}, our work on transferability on USENIX \cite{demontis19-usenix})

%Work on reprogramming for transfer learning (non-adversarial use) \cite{tsai20-pmlr}.

\section{Contributions, Limitations and Future Work}\label{sec:conclusions}
%Recap main contributions, as listed in the introduction.
%Sketch limitations/open issues with this work and future research directions; e.g., we may discuss how to extend our model to account for different reprogramming perturbations or to investigate the impact of universal perturbations.

Adversarial reprogramming has been originally proposed as an attack aimed to abuse machine-learning models provided as a service.
However, it has been subsequently shown that such a technique may also be used beneficially, providing a valuable approach to transfer learning. Despite its great practical relevance, no previous work has explained the main factors affecting the performance of adversarial reprogramming, \ie, why it works, when it fails, and how to tell the difference.

In this work, we have overcome this limitation by providing a first-order linear analysis of adversarial reprogramming, which sheds light on the underlying factors influencing reprogramming accuracy. We have then performed an extensive experimental analysis involving different machine-learning models and fourteen different reprogramming tasks. Thanks to our theoretical and empirical analyses, we have shown that the success of reprogramming depends on the size of the average input gradient, which is larger when the input gradients are more aligned, and when inputs have higher dimensionality. Our work thus provides a first concrete step towards analyzing the success and failure of adversarial reprogramming, paving the way to future work that may enable the development of better defenses against adversarial reprogramming, and improved transfer-learning algorithms. An interesting future development of this work also includes deriving guidelines to help practitioners to select machine-learning models which are easier to repurpose for a different task, thereby facilitating the process of transfer learning.

Two limitations currently exist in our work. 
The first is that it is not immediately clear from our analysis whether and to which extent the number of source- and target-domain classes may have an impact on the performance of adversarial reprogramming, and this aspect certainly deserves a more detailed empirical investigation in the future.
The second limitation is that we have only considered adversarial programs optimized within an $l_\infty$-norm constraint in this work. Nevertheless, our analysis can be easily extended to other $l_p$-norm perturbation models, as discussed in Sect.~\ref{sect:extension}, and it can also be exploited to provide deeper insights on attacks in which the adversarial perturbation is averaged over different input samples (including universal adversarial perturbations and robust physical-world adversarial examples), as discussed in Sect.~\ref{sect:rel-first-order}. We firmly believe that exploring these research directions will certainly provide a promising avenue for future work.

\section{Acknowledgements}
This work was partly supported by the PRIN 2017 project RexLearn, funded by the Italian Ministry of Education, University and Research (grant no. 2017TWNMH2); by BMK, BMDW, and the Province of Upper Austria in the frame of the COMET Programme managed by FFG in the COMET Module S3AI; and by the Key Research and Development Program of Shaanxi (Program Nos. 2022ZDLGY06-07, 2021ZDLGY15-01, 2021ZDLGY09-04 and 2021GY-004), the International Science and Technology Cooperation Research Project of Shenzhen (GJHZ20200731095204013), the National Natural Science Foundation of China (Grant No. 61772419).

%\balance
%\bibliography{addref,bibDB}

\end{document}